\documentclass[letter,12pt]{article}
\usepackage[margin=25mm]{geometry}

\usepackage{amsmath}
\usepackage{amssymb}
\usepackage{cite} 
\usepackage{graphicx}
\usepackage{caption}
\usepackage[hidelinks]{hyperref} 
\usepackage{cleveref} 
\usepackage[linesnumbered,ruled,spanish,onelanguage]{algorithm2e}
\usepackage{algpseudocode}

\title{Aligning a medium-size GPT model in English to a small closed domain in Spanish}

\author{Oscar R. Navarrete-Parra, Víctor Uc-Cetina, Jorge Reyes-Magaña \\
\small oscarranavpa@gmail.com, \small uccetina@correo.uady.mx, \small jorge.reyes@correo.uady.mx \\ \\
Facultad de Matemáticas \\
Universidad Aut\'onoma de Yucat\'an \\
}

\date{}

\begin{document}

\maketitle

\begin{abstract}
In this paper, we propose a methodology to align a medium-sized GPT model, originally trained in English for an open domain, to a small closed domain in Spanish. The application for which the model is finely tuned is the question answering task. To achieve this we also needed to train and implement another neural network (which we called the reward model) that could score and determine whether an answer is appropriate for a given question. This component served to improve the decoding and generation of the answers of the system. Numerical metrics such as BLEU and perplexity were used to evaluate the model, and human judgment was also used to compare the decoding technique with others. Finally, the results favored the proposed method, and it was determined that it is feasible to use a reward model to align the generation of responses.
\end{abstract}

\maketitle

\section{Introduction}
Transformer neural networks have shown great potential for natural language processing \cite{vaswani2017attention}. Several pre-trained transformer language models have been developed with the use of transfer learning, for instance the work by Radford et al.  \cite{radford2018improving}. These models have been scaled by increasing their number of parameters by hundreds of millions like GPT-2 \cite{radford2019language} and BERT \cite{devlin2019bert}, or up to hundreds of billions like GPT-3 \cite{brown2020language} and GPT-4 \cite{openai2023gpt4}. Transformer models are trained with massive amounts of text and once they have been finely aligned, they can even recognize the task they must perform, despite never having been specifically trained for it. It is also possible to use the weights of these models and fine-tune them for a particular task, for example, open domain conversational systems like DialoGPT \cite{zhang2020dialogpt}, InstructGPT and ChatGPT \cite{ouyang2022training}.

On one hand, training huge models like GPT-4 is complicated and highly computationally expensive. On the other hand, medium-size models still lack coherence and sometimes tend to hallucinate facts. Furthermore, since they are usually trained with a wide variety of texts on the Internet, their behavior is not aligned to correctly follow the user's instructions. This highlights the need for some additional component capable of providing consistency to the dialogue state, as well as some kind of system that evaluates and determines the quality of  the responses \cite{mctear2020conversational}. However, it is not always possible to have a large enough labeled dataset to fine-tune these models. Especially if the data comes from a knowledge base about a particular product, service or subject. Therefore, there is a need to search for feasible strategies for training language models in the response-generation task, that can be customized for a specific information domain.

In this article, we focus on the problem of aligning the DialoGPT model, to enable it to answer questions in Spanish. DialoGPT is a model based on GPT-2 and it was originally trained in English. In order to achieve a good alignment, the model is firstly fine-tuned with a corpus in Spanish. Then, it is further refined with a small collection of questions and answers about a specific topic. Finally, the performance of the model is improved through a reward neural network trained with human feedback. As a case study, we implemented a chatbot that answers frequently asked questions from a University undergraduate program, which we will refer to as the LCC dataset. The results show that it is feasible to use a reward model to align a medium-size GPT model to a small set of question-answer pairs.

\section{Rewards based on human preferences}
When modeling the environment for a reinforcement learning algorithm, a common practice is to manually design the reward function $R: S \rightarrow \Re$. However, for language processing, it is not as straightforward to design a reward system by hand. This is because the quality and accuracy of the generated sentences do not depend solely on their spelling and grammar, but also on their semantics and consistency with the query or assigned instruction. Therefore, to refine a language model through reinforcement learning, we need to employ a reward function that can distinguish the quality of the generated sentences, just as a person would. One way to achieve this is by training a reward model based on human preferences, that is, to train a sentence classifier in a supervised manner where the dataset is created from examples previously selected and labeled by human judges. Ziegler et al. \cite{ziegler2019finetuning} used this approach to fine-tune GPT-2, so that it could generate more realistic summaries. In order to do so, they represented the probability distribution of their model as a policy $\pi: S \times A \rightarrow [0,1]$, and trained a reward model using human taggers, choosing the best tag from a small set of possibilities, to subsequently train the policy with the reward function based on human preferences -- i.e. the rewards model learns to rate sentences like a human. One of the advantages of fine-tuning a language model with reinforcement learning instead of supervised learning is that such fine adjustment of the model requires less labeled data.

All of this suggests that it is possible to fine-tune pre-trained, not-too-giant language models by combining reinforcement learning techniques and human preferences, obtaining models capable of understanding a wider range of language and responding more accurately to user queries. In other words, language models can be practically aligned using reinforcement learning components such as a reward function.

\section{Methodology}

\begin{figure*}
\centering
\includegraphics[width=15cm]{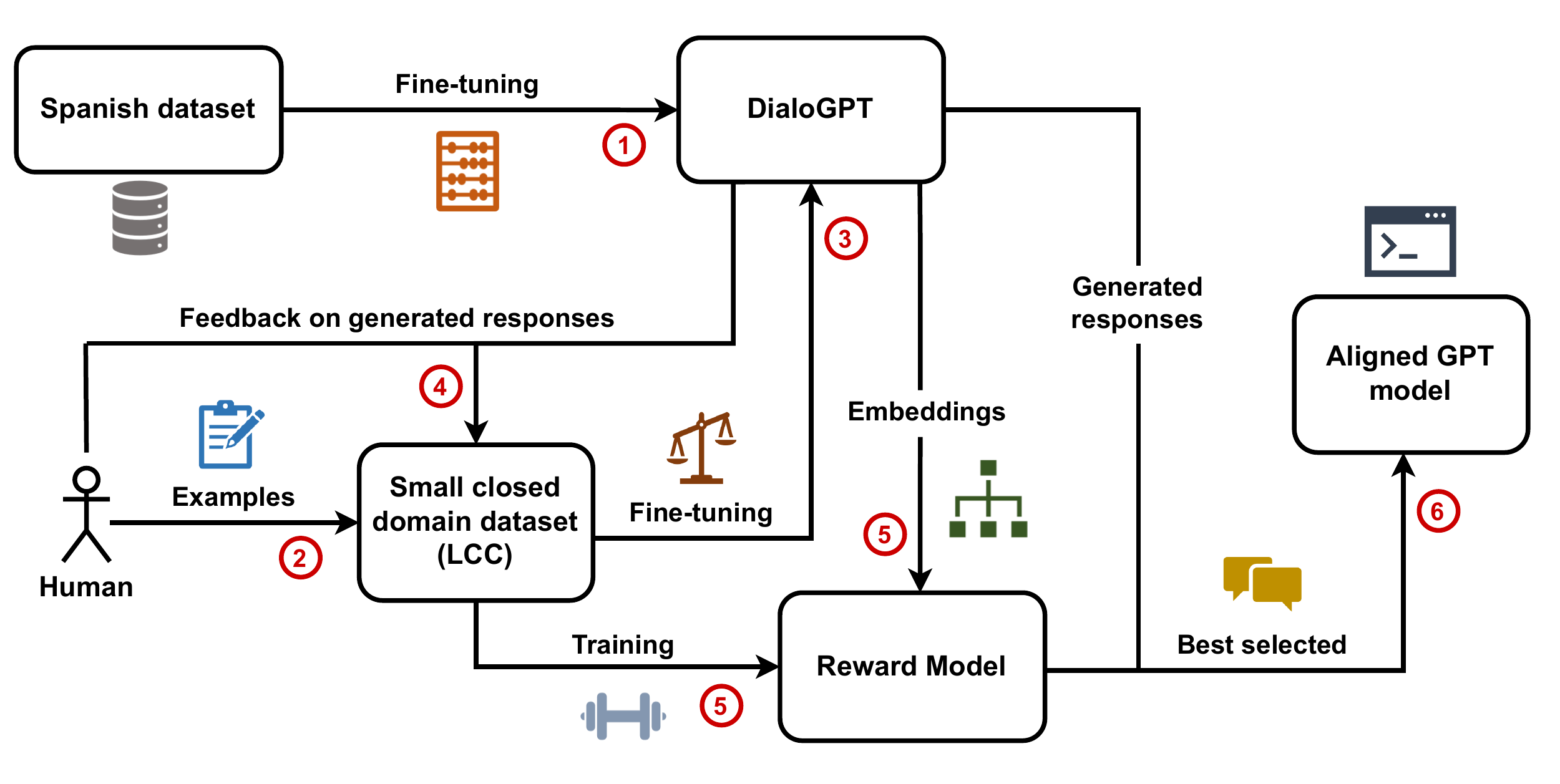}
\caption{Outline of the methodology.}
\label{fig:4_1}
\end{figure*}

The proposed methodology is illustrated in Figure \ref{fig:4_1}. In general we have the following components:

\begin{itemize}
\item Acquisition of a database in Spanish, which is used for the first fine-tuning of the DialoGPT conversational model. 

\item Acquisition of a small dataset about a particular topic, namely, the LCC dataset. 

\item Refinement of the conversational model using the LCC dataset.

\item Manual rating by humans of the new responses generated. The resulting rating is further used to augment the previous version of the dataset. 

\item The augmented dataset and the embeddings generated from the fine-tuned DialoGPT model are used to train a neural network that predicts rewards for each response, namely, the reward model.

\item Optimization of the conversational system via a decoding algorithm with rewards in order to align it to the specific-topic dataset.
\end{itemize}
	
\section{Conversational model}

In this work we use the DialoGPT model because it is a tunable and relatively small language model originally trained to generate dialogues in English. Moreover, it is possible to further refine this model and adapt it to a different language \cite{adewumi2021smaaprat}.

Thanks to the fact that DialoGPT is a sizable model, it is possible to train it without the need for huge and sophisticated computing equipment, unlike larger models such as GPT-3 and GPT-4. In this case, the medium-size DialoGPT model trained from GPT-2 checkpoint was used, and we further fine-tuned it using the Spanish conversation database. In this way, GPT-2's weights help the model to understand language structure while DialoGPT's weights provide conversational insight. The same idea is followed to adjust DialoGPT to the Spanish language, in order to further adapt it to the LCC dataset.

\subsection{Conversations dataset in Spanish}
To refine the model to the Spanish language, a subset of the original work database was translated. This dataset was obtained from the Reddit file server\footnote{File server address at \url{https: //files.pushshift.io/reddit/}.}, specifically, we extracted the conversations between the years 2006 and 2007 of the site. The resulting dataset had a size of 379KB and consisted of 110 thousand examples. We used OPUS-MT \cite{tiedemann2020opus} to translate the dataset from English to Spanish, employing the one-way translator model via the EasyNMT library\footnote{EasyNMT Github repository at the link \url{https://github.com/UKPLab/EasyNMT}.}. Prior to translation, we split the dataset into an $80\%$ training set and a $20\%$ validation set.
    
\section{Reward model}
The responses that the refined DialoGPT model generates are in Spanish but lack coherence and are not aligned to the user's intention. Furthermore, since the model was trained on a set of open domain conversations, the model does not really serve a useful purpose. To mitigate this problem, we used a reward model based on human preferences, in order to improve the quality of the responses generated by DialoGPT and align the model to a set of closed-domain dialogues focused on a particular topic.
    
\subsection{Creation of the dataset from human preferences}
Once the first fine-tuning on the Spanish dataset is completed, we proceeded to build the dataset from human preferences to further refine DialoGPT and to train a new reward model. This dataset was collected in two stages.

In the first stage, a human tagger creates a new set of (question, answer) pairs of a specific topic (i.e. the LCC dataset), generating between 50 and 100 examples, to slightly fine-tune the model. It is important to consider two factors during this stage. First, this aditional fine-tuning of the model must not drastically alter the current values of the parameters, otherwise, all previous training could be forgotten and the model would be limited to generating responses only from this dataset. On the other hand, it is at this moment that the dataset to which we want the model to be aligned with is included. So, more examples of dialogues can be added to improve its social behavior.

In the second stage, once the model has been slightly fine-tuned, the human tagger rates the newly generated responses on a scale of 0 to 1, using the dataset created in the first stage. We save the tuples (question, response, rating) into a new dataset that also contains examples from the previous set. This newly generated dataset is then used to train a reward model.

\subsection{Implementation of the reward model}
\begin{figure*}[ht]
\centering
\includegraphics[width=14cm]{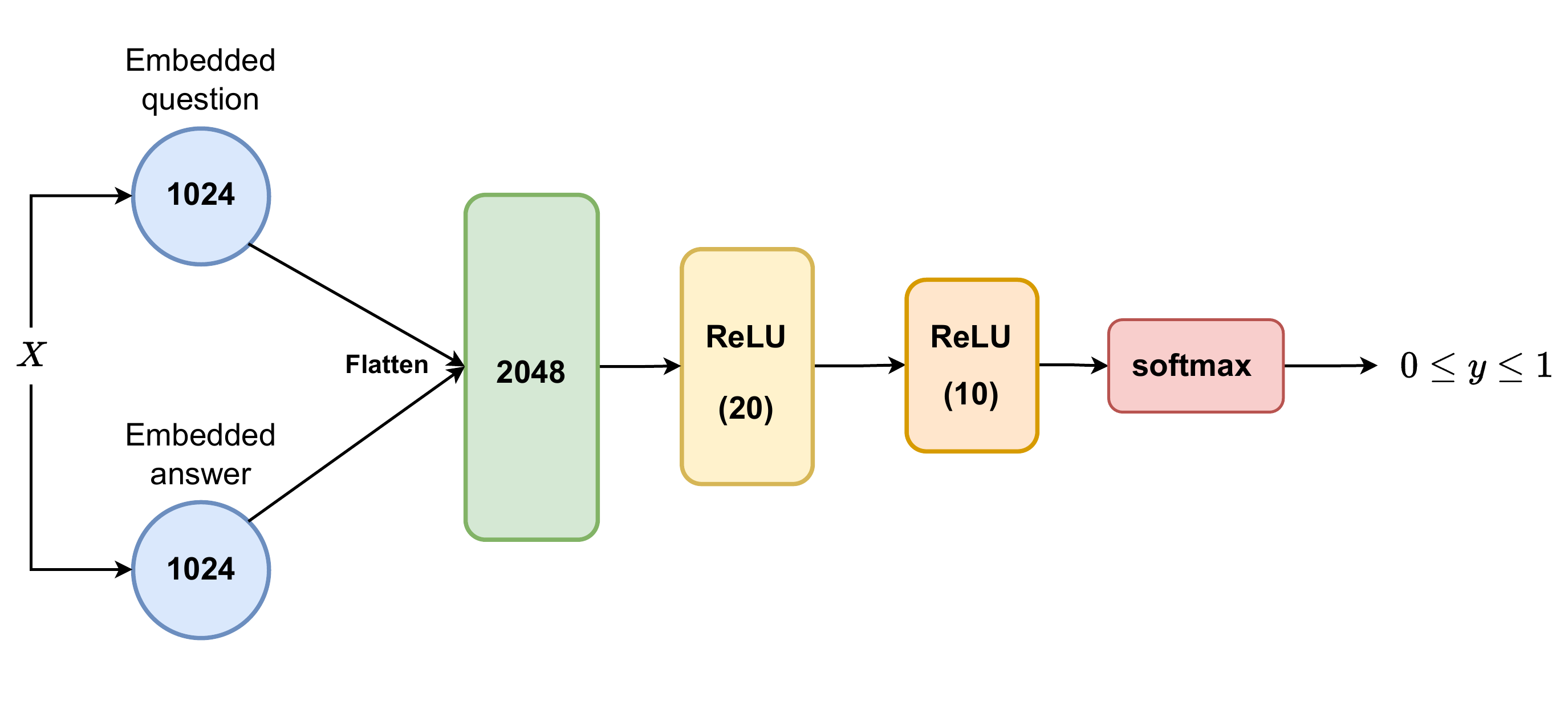}
\caption{Neural network architecture for the reward function.}
\label{fig:4_2}
\end{figure*}

The reward model is essentially a feed-forward neural network, and its architecture is described in Figure \ref{fig:4_2}. The network takes as input, the embeddings of two sentences, the question and its respective answer. Since the embedding layer of the fine-tuned model already provides a good representation of the Spanish words, this step speeds up the learning process. These vectors are then passed to a smoothing layer and the resulting vector is processed through two layers with ReLU activation function. In the final layer, the softmax activation function converts the values to probabilities, which will be used to directly qualify a (question, answer) pair.

\subsection{Optimization of the response generation}
The algorithm for optimizing the responses using the reward model is given in Algorithm \ref{alg:4_2}. First, we calculate $N$ candidates for the next token to be generated, based on the probabilities obtained from the language model. Then, we calculate the score for each of these candidates, using the already trained reward model, which takes as input the embeddings of the input sequence and the embedding of each candidate. The reward model assigns a score to each candidate, which measures its semantic similarity to the input sequence. Finally, the candidate with the highest score is selected as the next token to be generated.

The embeddings of a sentence are calculated by adding all the embeddings of each token appearing in that sentence. This process results in a fixed-length vector representation of the sentence that captures its semantic content.

    \SetKwComment{Comment}{/* }{ */}
    \SetKwFunction{Modelo}{Model}
    \SetKwFunction{Softmax}{softmax}
    \SetKwFunction{Multinomial}{multinomial}
    \SetKwFunction{Append}{Append}
    \SetKwFunction{Decode}{Tokenizer.decoder}
    \SetKwFunction{Tokenizer}{Tokenizer}
    \SetKwFunction{Reward}{Reward}
    \SetKwFunction{EOS}{.EOS}
    \SetKwFunction{Embedding}{Embedding}
    \SetKwFunction{Argmax}{ArgMax}
    \SetKwInOut{Input}{Input}
    \SetKwInOut{Output}{Output}
    \SetKw{KwBreak}{Break}
 
    \begin{algorithm*}[hbt!]
    \caption{Decoding algorithm with rewards}\label{alg:4_2}
    
    \Input{$X$: Input token sequence}
    \Input{$T$: Temperature}
    \Input{$L$: Maximum output length}
    \Input{$N$: Number of candidates}
    \Output{$Y$: Output text sentence}
    \BlankLine
    
    $Candidates \gets [ ]$ \Comment*[r]{$candidates$ is initialized as an empty list}
    \For{$i\leftarrow 1$ \KwTo $N$}{
        $Y \gets [ ]$\;
        \For{$j\leftarrow 1$ \KwTo $L$}{
          $logits \gets$ \Modelo{$X$}$[-1] / T$\;
          $token \gets$ \Multinomial{\Softmax{$logits$}}\;
          \Append{$token$, $Y$}\;
          \If{$token = \Modelo\EOS$}
          {
            \KwBreak\;
          }
          $X \gets X + token$\;
        }
        \Append{$Y$, $Candidates$}\Comment*[]{$Y$ is added to the candidate list}
    }
    $Scores \gets [ ]$\;
    \For{$i\leftarrow 1$ \KwTo $N$}{
    $c \gets $\Reward{\Embedding{$X$},\Embedding{$Candidates[i]$}}\;
    \Append{$c$, $Scores$}\;
    }
    $m \gets$ \Argmax{$Scores$}\;
    \Return \Decode{$Candidates[m]$}\;
    \end{algorithm*}

\section{Experimental work}
For the training of both the language and reward models, a computer with Intel(R) Xeon(R) 2.30GHz CPU, 12.7 GB RAM memory, Nvidia Tesla T4 GPU with 16GB GDDR5 1.59GHz and 107.7 GB of hard disk storage was used. In the first instance, the model was fine-tuned using the Spanish dataset, and later it was further refined using the LCC dataset. The hyperparameters used in these two training phases are shown in Table \ref{tab:5_1}. 

To prevent overfitting, we used regularization by penalty of L1 weights during training. In addition, we employed the Noam decay scheme \cite{vaswani2017attention} with a warmup of 200 steps, which dynamically adjusts the learning rate during training to improve convergence.

\begin{figure}
\centering
\includegraphics[width=8cm]{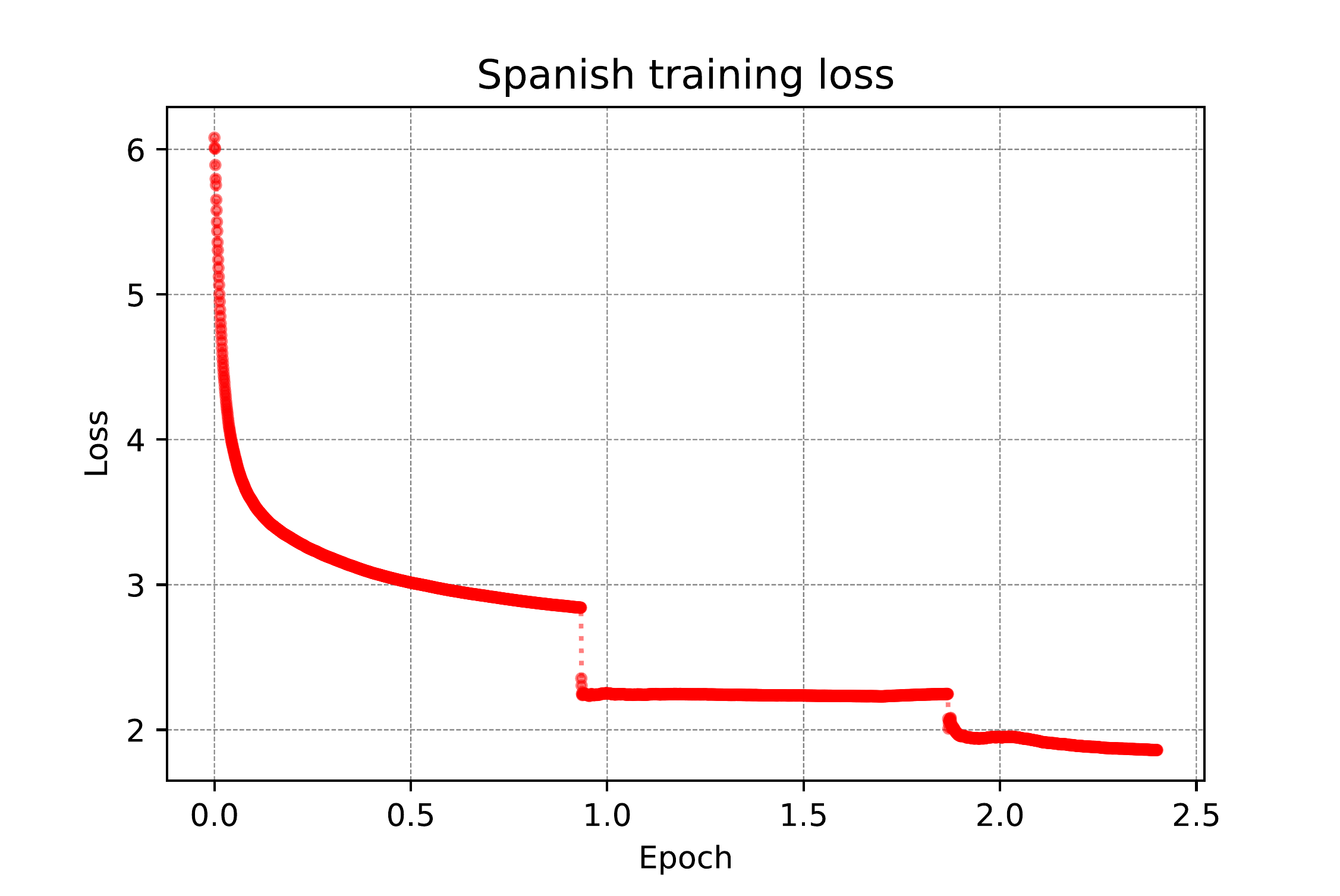}
\caption{Graph of the cost function for training with the dataset in Spanish.}
\label{fig:5_1}
\end{figure} 

\subsection{DialoGPT model training}

\begin{table*}
\centering
\begin{tabular}{|l|c|c|}
\hline
        & \multicolumn{1}{l|}{Spanish model} & \multicolumn{1}{l|}{Spanish model + LCC dataset} \\ \hline
Batch size    & 64                                     & 1                         \\ 
Learning rate      & $1\times10^{-5}$             & $1\times10^{-5}$                \\
Scheme & Noam decay                             & None                            \\
Epochs  & 2.5                                    & 2                                  \\
Reg. factor  & 0.01                                   & 0.001                       \\
Time  & 4 hours                                & 10 minutes                           \\ \hline

\end{tabular}
\caption{Training hyperparameters.}
\label{tab:5_1}
\end{table*}

\begin{figure}
\centering
\includegraphics[width=8cm]{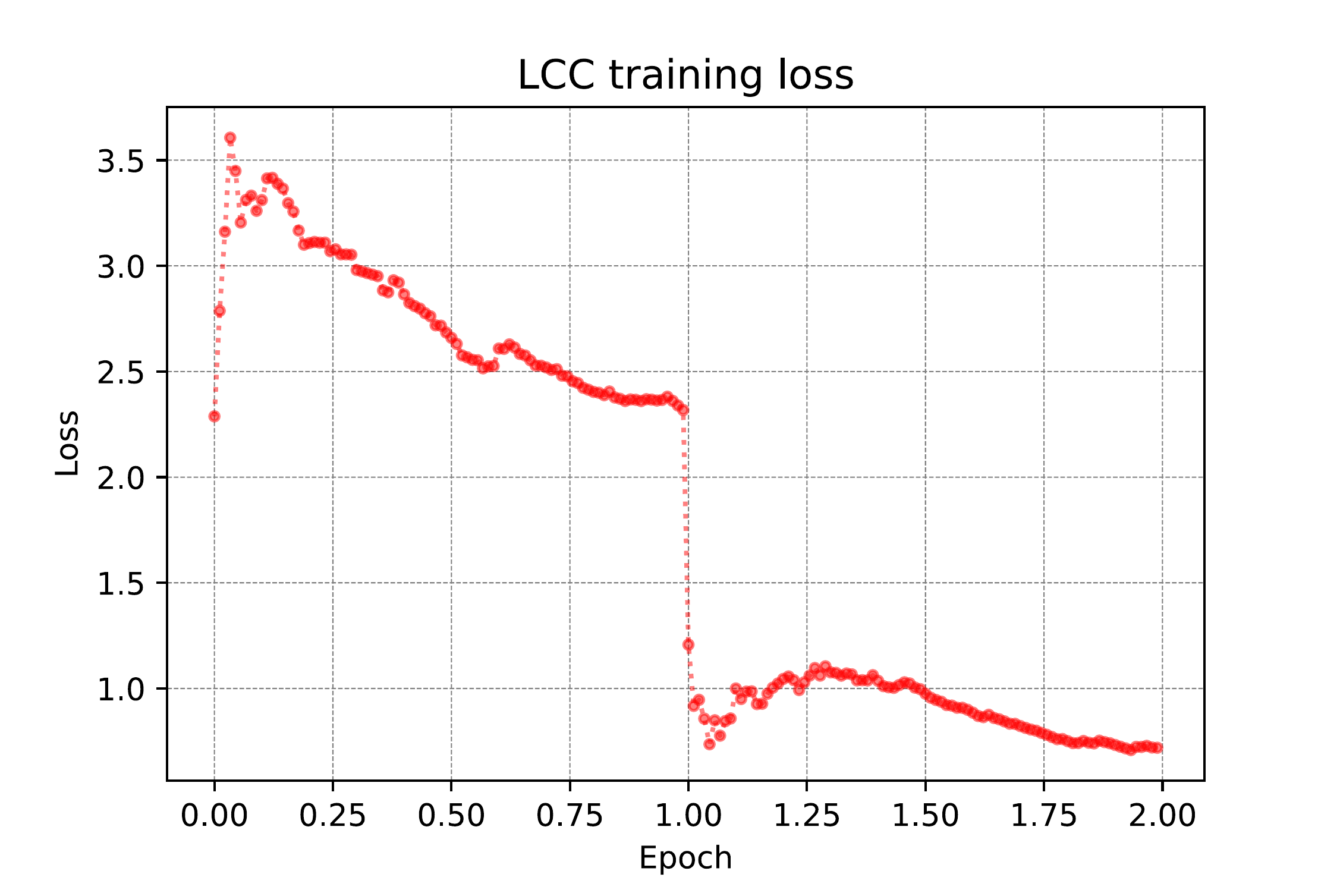}
\caption{Plot of the cost function for training with the LCC dataset.}
\label{fig:5_2}
\end{figure}

Figure \ref{fig:5_1} shows the result of the first training. Notice how the model converges rapidly during the first few iterations, and then it starts to slow down. This is because the model was not tuned from scratch, and although it was trained in another language, the model already had basic knowledge about conversations and language. During the second training, the loss value of the model is very unstable at the beginning, which is likely due to the fine-tuning of the model's pre-existing parameters. However, once the first epoch is completed, the model converges extraordinarily quickly (see Figure \ref{fig:5_2}). This may be due in part to the many-to-many structure of the LCC dataset, which contains a diverse range of questions and answers, allowing the model to learn more effectively from different types of conversational data.

\subsection{Reward model training}

\begin{figure}
\centering
\includegraphics[width=8cm]{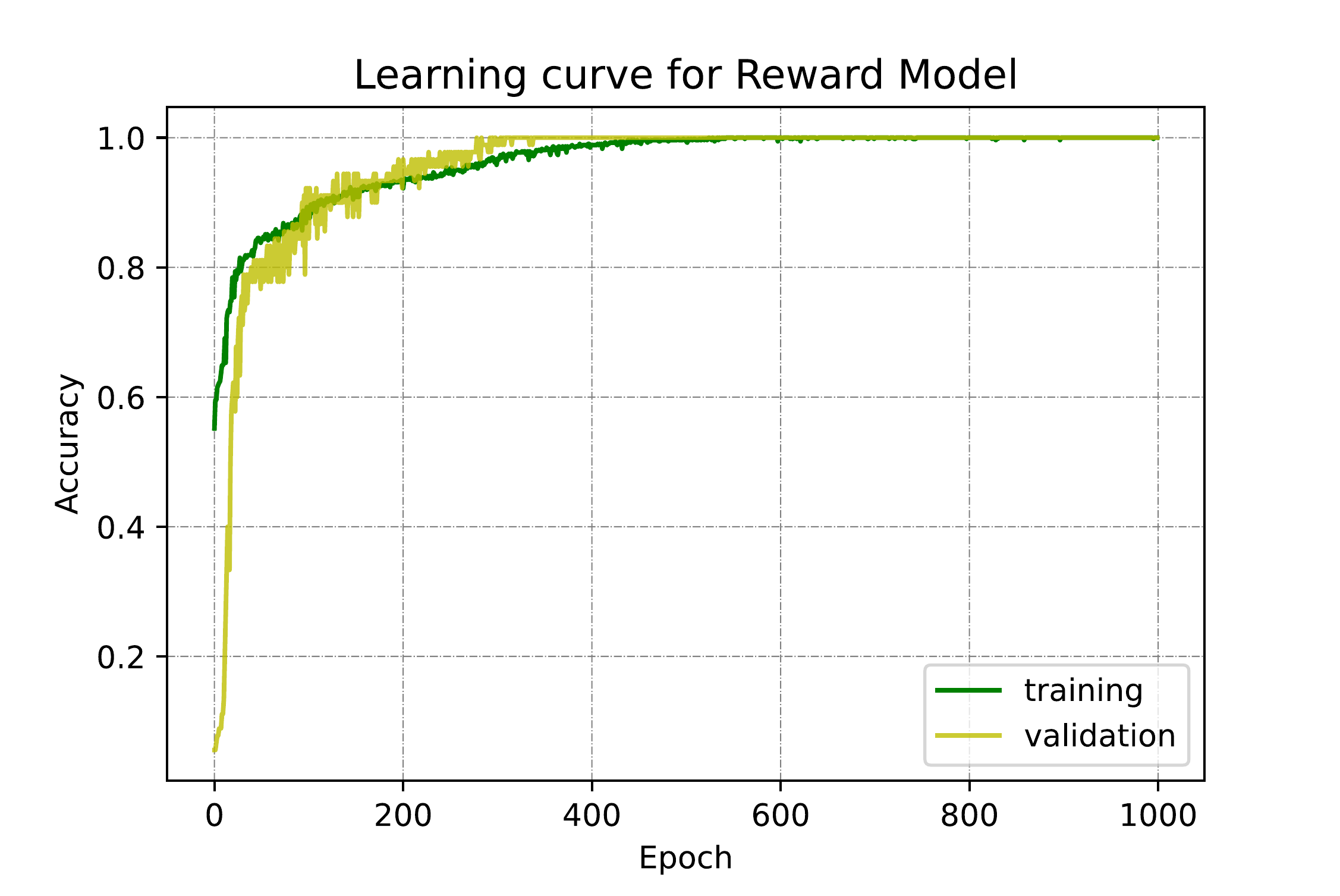}
\caption{Reward model learning curve graph.}
\label{fig:5_3}
\end{figure}

Using the LCC dataset and the model tuned with it, 5 answers were generated for each question using sampling. Then, each pair (question, answer) was labeled with a score in the interval $(0, 1)$. All 90 examples from the previous LCC dataset were added to this new set, with a score of 1.0, making a total of 540 (question, answer, score) examples. The resulting dataset was used to train the reward model for 1000 epochs, with a learning rate of 0.001. To evaluate this model, the precision of the neural network was measured in two datasets: the training and the validation one; the latter is the original LCC dataset, and was used to verify that at least the learned reward function is capable of giving a positive score to all correct (question, answer) pairs. Figure \ref{fig:5_3} illustrates both learning curves and shows that the model has learned to correctly reward all right answers for each question, however, it is still difficult to assess whether the model is capable of scoring all the wrong dialogues with a negative rating.

\section{Evaluation of the generated responses}

In order to evaluate the performance of the reward model, the generated responses were evaluated by comparing the system with 3 different decoding algorithms: filtered sampling with top-$k$$=20$ and top-$p$$=0.9$ , sampling with the same filtering as above and using the rewards model with 10 candidates, and beam search with $b=6$; we choose a generation length of $200$ tokens and temperature $T=1$.

For the automatic evaluation, the perplexity metric and the BLEU  score \cite{papineni2002bleu} were used. Perplexity measures how well the model's output probability distributions predict the target tokens, therefore a low perplexity corresponds to a better performance. While BLEU analyzes if the generated sentences contain sets of occurrences that are also present in the training examples.

\subsection{Results and discussion}

\begin{table*}
\centering
\begin{tabular}{|l|c|c|c|}
\hline
    & \multicolumn{1}{l|}{Sampling} & \multicolumn{1}{l|}{Sampling with reward} & \multicolumn{1}{l|}{Beam search} \\ \hline
BLEU    & 20.70                         & 16.99                           & 7.82                                      \\ \hline
Perp. & 8.49                          & 6.20                            & 12.09                                     \\ \hline
\end{tabular}
\caption{Evaluation with numerical metrics}
\label{tab:5_2}
\end{table*}

\begin{figure*}
\centering
\includegraphics[width=12cm]{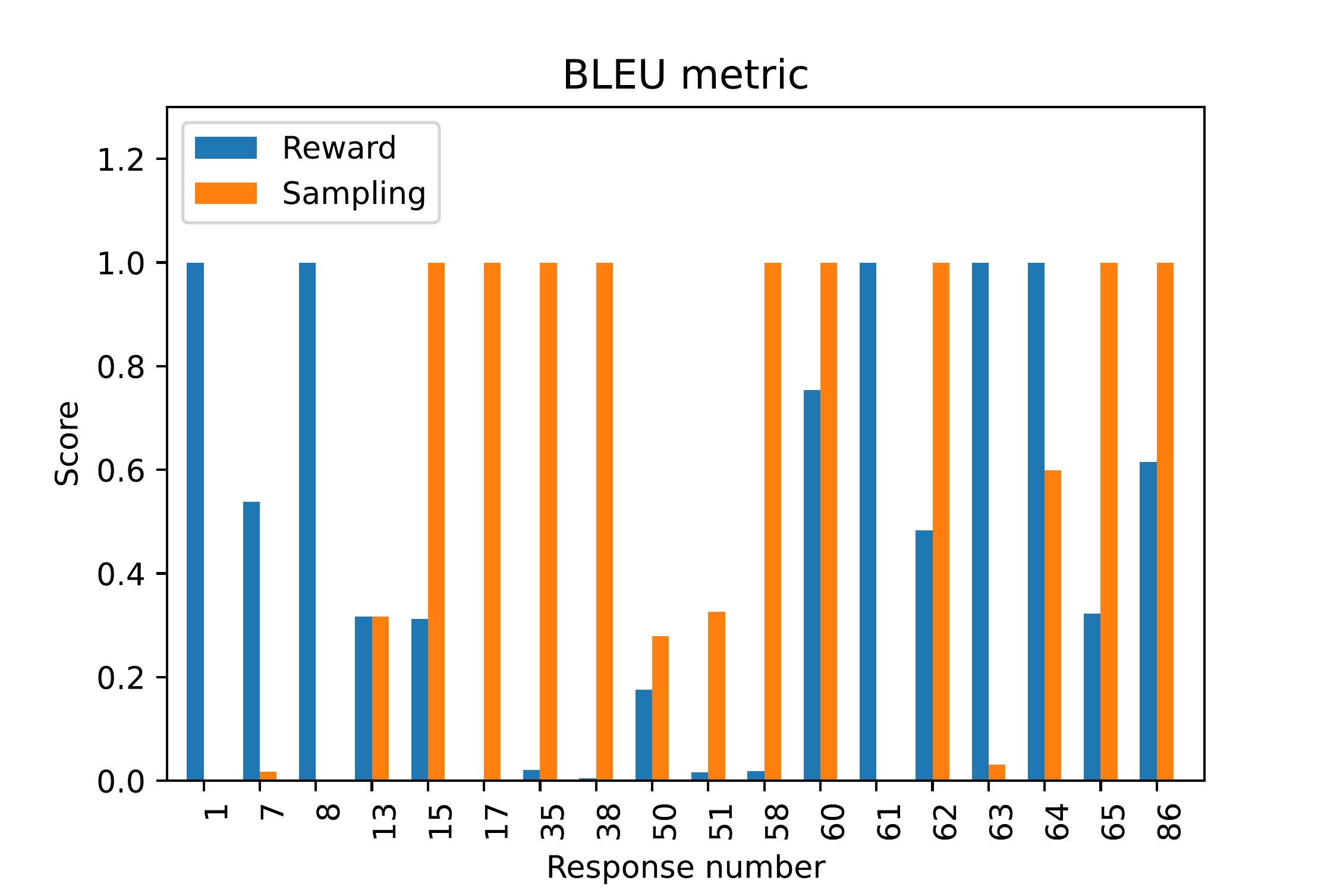}
\caption{Evaluation using BLEU.}
\label{fig:5_4}
\end{figure*}   

\begin{figure*}
\centering
\includegraphics[width=12cm]{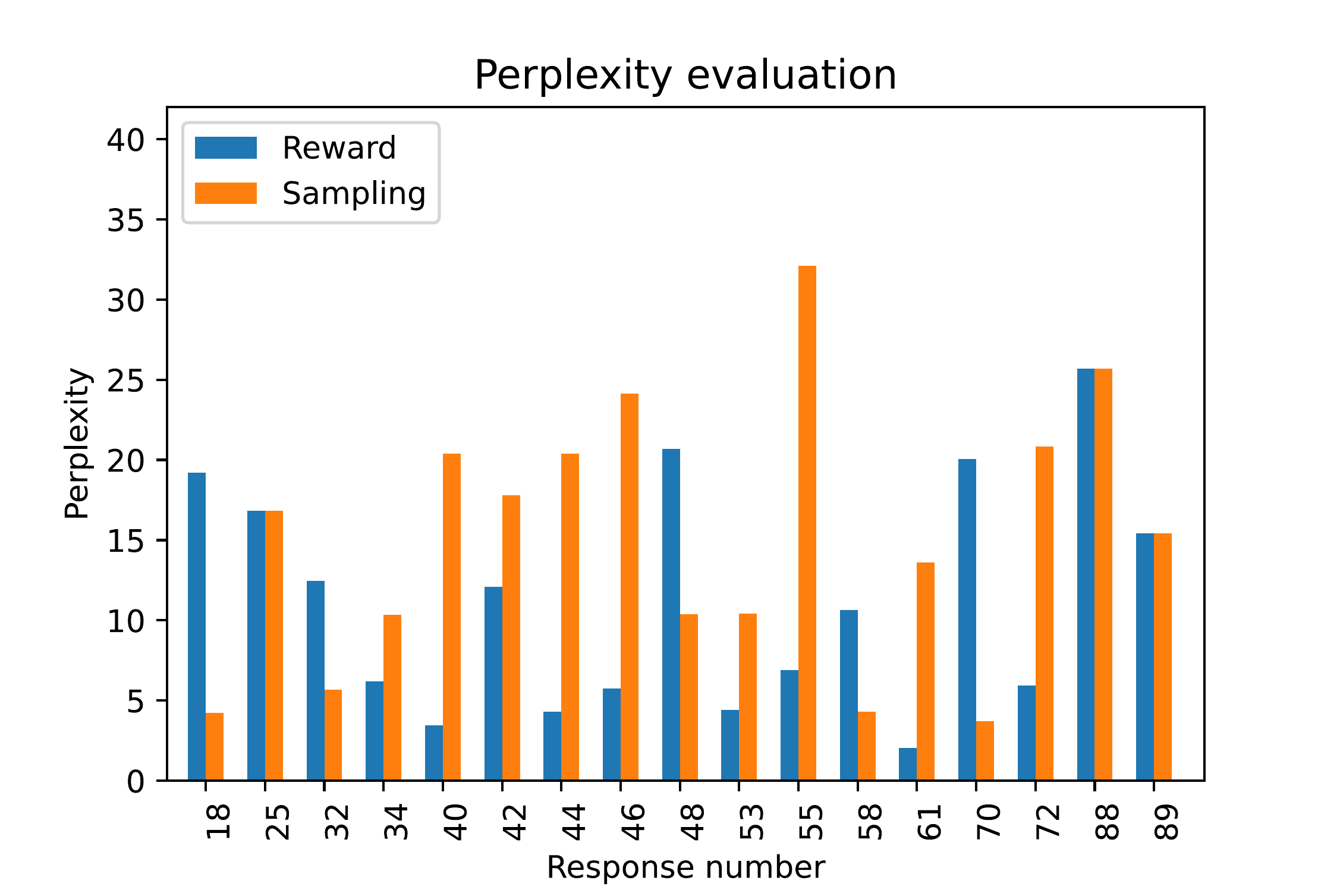}
\caption{Evaluation using perplexity.}
\label{fig:5_5}
\end{figure*}  

Table \ref{tab:5_2} shows the results.  Using the BLEU metric, normal sampling achieved a higher accumulated score on the 90 instances while reward sampling obtained almost 4 points less. Both methods should have received a score of at least 45 as almost half of the predicted responses were correct. However, this was not the case because BLEU automatically penalizes with 0 those very small $N$-grams such as uni-grams, bi-grams and tri-grams. Although the sampling technique was superior in this evaluation, this does not mean that it is better due to the lack of references and the deficiency of evaluating a conversational system by comparing $N$-grams, because unlike machine translation, a question can have too many associated answers.

On the other hand, in the evaluation of perplexity, sampling with rewards showed greater superiority, this means that the rewards model is capable of generating a better probability distribution than simple sampling. Furthermore, both techniques significantly outperformed the traditional beam search technique, which indicates that response generation is a more complex task than simply maximizing token sequence probabilities.

Likewise, an evaluation of human judgment was also carried out. That is, the answers that each technique generated through the review of a person were evaluated, considering characteristics of the answers such as coherence and relevance with respect to the question. The outcome was the preference towards the responses generated by the reward sampling technique. Some examples of conversations generated using reward sampling are shown in Table \ref{tab:5_3}.

About 44\% of the responses that the model generated using this technique were either identical to those of the original dataset or at least appropriate according to the user's intent. Another 32\% of the responses did not provided true information but instead hallucinated facts or the model wasn't able to correctly interpret the query. The rest of the responses were basically gibberish incoherent sentences. Many of these examples were ridiculously complex explanations and made no semantic sense, while others tended to have many grammatical errors with nonsensical word repetitions and made-up conjugations. This is mostly because the model was trained on a very small dataset. Language models typically tend to be trained with millions of instances to allow the model to fully capture the meaning of phrases and words. Another reason why the chatbot, even using the reward model, generates these types of responses may be due to the quality of translation from English to Spanish of the first dataset. In this case, it was not possible to manually review and correct the translation of each sentence to avoid this inconvenience. Despite this, the technique implemented with the reward model generated a larger number of correct responses and fewer gibberish responses, compared to sampling without rewards.

\begin{table*}
\centering
\begin{tabular}{|p{0.35\linewidth} | p{0.6\linewidth}|}
\hline
User & Chatbot \\
\hline
¿En qué áreas me puedo especializar? & Inteligencia Artificial, Cómputo Científico, Ciencias de Datos, Redes de Computadoras, Sistemas Operativos, Realidad Aumentada, entre otros.\\ \hline
¿Tengo que estudiar mucho? & En esta carrera se presentan las bases matemáticas que dan paso a las aplicaciones de la computación. También se da una introducción a la algoritmia y programación. Es cuestión de práctica. Si te gustan las matemáticas y lógica, la programación y los retos ¡te recomendamos Ciencias de la Computación!\\ 
\hline
¿Cuántos años se estudia? & El plan actual tiene una duración desde 9 semestres a 18 semestres.\\ 
\hline
¿Hay apoyos económicos? & Sí, hay varios tipos de apoyos y becas que puedes consultar en los sitios oficiales.\\ 
\hline
¿Qué lenguajes de programación aprenderé? & Te enseñamos desde lo básico y también varios tipos de lenguajes de programación. Aquí aprendemos a programar en lenguaje C, Java y Python.\\ 
\hline
\end{tabular}
\caption{Examples of conversations using reward sampling.}
\label{tab:5_3}
\end{table*}

\section{Conclusion}
In this work, a pre-trained state-of-the-art transformer model for dialogue generation was implemented and refined through two consecutive fine-tuning phases in order to align it to a small dataset of questions and answers about one specific undergraduate program at our University. To this end we used two datasets. One dataset was used to fit the model to the target language and the second one was used to align it to the closed-domain dataset. For this second step, a separate neural network was also implemented and trained, so that it learns to qualify and reward conversations according to human preferences. This model was trained with feedback data from the conversational system. Finally, the reward model was used to optimize the dialogue generation of the tuned transformer and its performance was compared with two other response decoding techniques. The result was a decent improvement over the filtered sampling technique and a huge superiority over beam search. The results also show that it is feasible to align a medium-sized language model to a small dataset for a complex problem such as response generation.

Finally, it is important to mention that the use of a corpus originally generated in Spanish, and not an automatically translated one, as we did in this work for the first fine-tuning of the conversational system, would have certainly achieved a positive impact in the final performance. Also, the use of a more recent language model, like a larger GPT version, would improve the quality of the responses in the question-answer application, considering of course that retraining any of the newest data-hungry versions would impose some technical constraints, in terms of the computational power needed to do it.

\bibliographystyle{acm}
\bibliography{biblio}

\end{document}